# Learning Time Reduction Using Warm Start Methods for a Reinforcement Learning Based Supervisory Control in Hybrid Electric Vehicle Applications

Bin Xu[1,*], Jun Hou[2,*], Junzhe Shi[3], Huayi Li[4], Dhruvang Rathod[1], Zhe Wang[1], Zoran Filipi[1]

[1]: Clemson University, Department of Automotive Engineering, 4 Research Dr., Greenville, SC 29607, USA.

[2]: University of Michigan, Department of Electrical Engineering and Computer Science, 1301 Beal Avenue, Ann Arbor, MI, 48109, USA.

[3]: University of California, Berkeley, Civil and Environmental Engineering Department, 760 Davis Hall, Berkeley, CA 94720, USA.

[4]: University of Michigan, Department of Mechanical Engineering, 2350 Hayward St, Ann Arbor, MI 48109, USA.

* Corresponding author: Bin Xu, Clemson University, Department of Automotive Engineering, 4 Research Dr., Greenville, SC, 29607, USA. Phone: 864-626-8335, Email: xbin@clemson.edu. Jun Hou, University of Michigan, Department of Electrical Engineering and Computer Science, 1301 Beal Avenue, Ann Arbor, MI, 48109, USA. Phone: 734-272-5516, Email: junhou@umich.edu.

***Abstract*—Reinforcement Learning (RL) is widely utilized in the field of robotics, and as such, it is gradually being implemented in the Hybrid Electric Vehicle (HEV) supervisory control. Even though RL exhibits excellent performance in terms of fuel consumption minimization in simulation, the large learning iteration number needs a long learning time, making it hardly applicable in real-world vehicles. In addition, the fuel consumption of initial learning phases is much worse than baseline controls. This study aims to reduce the learning iterations of Q-learning in HEV application and improve fuel consumption in initial learning phases utilizing warm start methods. Different from previous studies, which initiated Q-learning with zero or random Q values, this study initiates the Q-learning with different supervisory controls (i.e., Equivalent Consumption Minimization Strategy control and heuristic control), and detailed analysis is given. The results show that the proposed warm start Q-learning requires 68.8% fewer iterations than cold start Q-learning. The trained Q-learning is validated in two different driving cycles, and the results show 10-16% MPG improvement when compared to Equivalent Consumption Minimization Strategy control. Furthermore, real-time feasibility is analyzed, and the guidance of vehicle implementation is provided. The results of this study can be used to facilitate the deployment of RL in vehicle supervisory control applications.**

## I. Introduction

With the strict fuel economy and emission regulations, OEMs and researchers are searching for propulsion system efficiency improvement techniques for passenger vehicles. Propulsion system electrification, as one of the efficiency improvement techniques, arose in the last two decades [1-3]. Hybrid Electric Vehicle (HEV) is an important step towards the vehicle propulsion system electrification as it directly improves the engine operating efficiency by using electric motor/ generator (EM) to help drive the vehicle [4].

Supervisory control is the brain of the HEV as it decides the operating conditions of the engine, EM, and battery. Thus, it has a significant impact on the HEV performance. Numerous HEV supervisory controls were researched in the past two decades, and the popular ones include rule-based method [5], Dynamic Programming (DP) [6], Equivalent Consumption Minimization Strategy (ECMS) [7], and Model Predictive Control (MPC) [8]. Rule-based method can be directly implemented in real-time as it has the lowest computation cost among four supervisory controls. However, the heuristic rules from experts' expertise or experiments might not be optimal, which leads to the need for global optimization, such as DP, to optimize the rules. DP generates global optimal results offline. Then, the rules are extracted from DP results and implemented in real-time. The DP assisted rule-based method is regarded as an indirect optimization method because the optimization is conducted offline by DP. To move the optimization from offline to real-time, ECMS and MPC supervisory controls are investigated. ECMS minimizes the instantaneous equivalent fuel consumption of engine and EM at each time step based on the given vehicle speed and torque demand. Compared to ECMS, MPC optimizes the fuel consumption and other HEV metrics in a future time window, leading to a higher computational time. However, both ECMS and MPC cannot guarantee global optimization solutions. Compared to the supervisory controls discussed above, reinforcement learning (RL) as another supervisory control has several advantages: (1) Compared with the rule-based method, RL is an optimization-based method and can achieve better fuel economy; (2) Compared with DP, RL requires much less computation cost and is capable of running online; (3) Compared with ECMS and MPC, RL is model-free and thus is not constrained by the reduced order model accuracy. Furthermore, ECMS only optimizes the fuel consumption at the current time step, and MPC only optimizes the fuel consumption for a short horizon, while RL is not limited by the instant or short time window and aims to find the global optimum with the help of Bellman equation [9].

In the past five years, RL-based HEV supervisory control gained more and more researchers' attention. RL mimics human learning behavior in a new environment by interacting with the environment. It explores the environment by trying different actions and receiving state and reward information from the environment. Actions in supervisory control are generally torque or power split between engine and EM for parallel HEV [10], and EM torque, engine speed/ torque for series HEV [11]. Some common states are vehicle speed, vehicle acceleration, vehicle power, battery SOC [12]. RL methods have shown successful performance in terms of energy-saving or fuel economy when compared to existing controls like heuristic, rule-based, ECMS, and DP. Deep RL was implemented as the power split algorithm in a hybrid electric bus [13]. Neural networks were used to approximate the Q values, and deep Q-learning achieved 89% of DP fuel economy. A hierarchical table-based RL framework was proposed to save energy for a plug-in fuel cell HEV [14]. There was no internal combustion engine in the propulsion system, and only fuel cell and lithium battery were considered. The higher level RL determined the penalty coefficient of fuel cell to avoid frequent fuel cell start or stop. The lower level RL determined the power split between fuel cell and lithium battery, based on the real-time penalty coefficient determined by the higher level RL. The proposed framework consumed 8.65% more hydrogen compared with DP. A Q-learning strategy was implemented in an HEV to split power among fuel cell, lithium battery, and ultracapacitor [15]. Compared with ECMS strategy, the proposed Q-learning RL strategy achieved 7-10% fuel saving in city and highway driving cycles. A table-based Q-learning was implemented in a plug-in HEV [16]. The proposed table-based Q-learning saved 16.8% energy compared with a rule-based energy management strategy. A deep RL algorithm was proposed to control a Plug-in HEV [17]. 16.3% energy saving was achieved by the deep RL using rule-based control as the baseline strategy. A model-based Q learning strategy was proposed for a parallel HEV [18]. The model was online trained via the interaction with the vehicle. More than 10% of fuel saving was achieved at city and highway driving cycles when compared with the rule-based method. An inner-outer loop deep RL framework was proposed to plan the path and save energy for a Toyota Prius plug-in HEV [19]. The outer loop generated the difference between the existing operation and the best operation recorded in history. The inner loop was an actor-critic RL framework.

Though RL-based HEV supervisory control gains momentum in recent years and shows encouraging energy-saving performance compared with existing controls, it has one big obstacle in the path of real-world implementation, which is its long learning time. Large number of iterations are required to achieve convergence of the RL-based supervisory control when neural networks are used to approximate Q values, such as 700-1,200 iterations [20], 15,000 iterations [13], 300,000 iterations [16], 150, 000 iterations [17], and 200,000 iterations [21]. Among the above RL literature in HEV applications, some used table-based RL methods [16, 20, 21], and the iterations varied from 700 to 300,000. The others used neural network based RL methods [13, 17], and the iterations varied from 15,000 to 150,000. This long learning time obstacle exists not only in HEV application, but also in most applications using RL algorithms. In [22], the study investigated multiple robotics tasks, and the RL algorithm required 100,000 iterations to reach convergence. In [23], a door open task was executed by a robotic arm. The RL algorithm in simulation required over 5,000 iterations to reach convergence, whereas, in experiments, the RL algorithm took more than 50,000 iterations to reach convergence. The above literature showed the RL-based supervisory control required more than thousands of iterations in the learning process to achieve satisfactory performance, and some even required more than 100,000 iterations. This large number of iterations require months of road or chassis dyno tests, which are costly both in time and money.

To reduce the learning time of the RL algorithms, several techniques are available: (i) learning rate adjustment [24], (ii) selective learning experience [25], and (iii) warm start [26]. Both the learning rate adjustment and selective learning experience have limited impacts on the learning time reduction due to the trade-off between learning time and vehicle performance. However, a warm start is not limited by that trade-off as it directly gives a one-time boost of vehicle performance in the initialization of the RL algorithm, where both the learning time and vehicle performance are improved at the same time. There is no warm start focused study in the field of HEV supervisory control, and only one study used a warm start method to facilitate learning convergence [20]. This reference used near-optimal Q values in the initialization, and the vehicle platform is plug-in HEV. It took 1200 driving cycles to converge for the cold start method and 700 driving cycles to converge for the warm start method. The difference between the reference and our work can be summarized as: (1) Our vehicle is SOC sustaining HEV rather than plug-in HEV (i.e., A plug-in HEV has more batteries than a SOC sustaining HEV does); (2) Our research provides surficent analysis as well as the detailed explanations of the warm start process. In the reference, only one sentence is given to describe the warm start and results (i.e., the initialization with simulated optimal or near-optimal solutions helps achieve a faster convergence); (3) It is not clear how the warm start process is achieved. Beyond the HEV application, warm start RL is researched in healthcare [26] and robotics [27]. Besides RL applications, warm start was also utilized in supervised learning [28, 29]. Overall, the warm start focused works are still lacking in RL, and, to our best knowledge, no study is found in HEV supervisory control. Additionally, in the vehicle warm start application, the engine and EM efficiency maps and the rules from experts are different from robotics and healthcare applications. More studies are required to investigate the RL-based supervisory control using warm start.

This study aims to investigate the warm start of RL based supervisory control in HEV application. To the best of our knowledge, this work has never been done before. The main contributions of this study are listed below:

(1). This study shows the learning time reduction advantage of warm start Q-learning over cold start in HEV application.

(2). For the first time, it achieves warm start Q-learning using two supervisory controls: ECMS and heuristic control.

(3). For the first time, it presents the detailed analysis of warm start and cold start comparison using RL state-action value function map, engine/ EM operating efficiency map, torque, speed, battery SOC, engine fuel consumption.

(4). It compares the warm start Q-learning supervisory control with ECMS control both in the driving cycle used in learning and the other two driving cycles Q-learning not used in learning.

The rest of the paper is organized as follows: Section II introduces the vehicle propulsion system model. Section III covers the Q-learning strategy and warm start strategies that utilized in this study. Reward, states, and actions are discussed in the Q-learning subsection. Heuristic strategy and ECMS are presented in the warm start strategies subsection. Section IV analyzes the results of cold start and warm start in detail and compares warm start Q-learning with ECMS in three different driving cycles (one cycle is the same one used in learning and two cycles that are not used in learning). Some key performance matrices are compared for the illustration, such as fuel consumption, engine/ EM torque, cycle speed tracking performance. In addition, real-time implementation feasibility of the warm start Q-learning is evaluated, and vehicle deployment guidance is given at the end of Section IV. Finally, the conclusion and future work are summarized in Section V.

## II. MODELING OF HEV PROPULSION SYSTEM

This study focuses on a middle-sized passenger car with a parallel hybrid propulsion system, and its architecture is shown in Fig. 1. Similar architecture can be found in BMW i8, which allocates an engine in the back and an electric motor in the front. The engine drives the front wheels, and the EM drives the rear wheels. The rated power of the engine is 64.6kW at 6000rpm, and the rated power of the EM is 143kW. There are four gear ratios in the transmission. The battery pack has 5.2kWh capacity. The primary parameters of the vehicle are summarized in Table 1. The vehicle model is built in Matlab/Simulink.

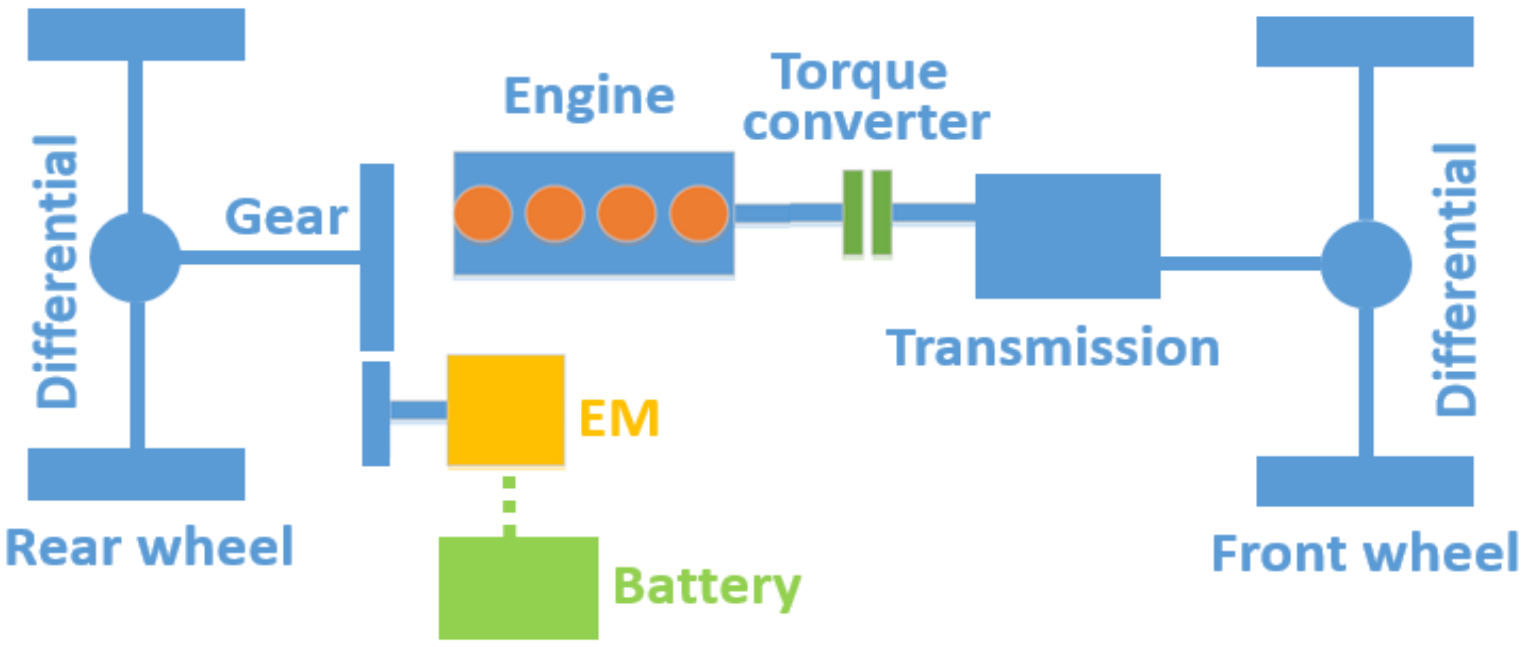

Fig. 1. Propulsion system architecture of the parallel HEV.

TABLE 1.

Parallel HEV specification.

| | |
|---|---|
| Curb weight | 1636 kg |
| Engine max torque | 115 Nm |
| Engine max power/ speed | 64.6kW @6000rpm |
| EM max torque | 400 Nm |
| EM max power | 143 kW |
| Transmission gear number | 4 |
| Transmission gear ratio | 1st/2.847, 2nd/1.552, 3rd/1.000, 4th:0.700 |
| final differential ratio | 4.13 |
| Battery cell number | 250 |
| number of series connection | 250 |
| number of parallel connection | 1 |
| Battery capacity (single cell) | 6.5 Ah |

### A. Torque Demand Modeling

The vehicle acceleration force is calculated as follows:

$$F_{acc} = F_{air} + F_{roll} + F_{grade} + F_{trac} \tag{1}$$

where $F_{air}, F_{roll}, F_{grade}, F_{trac}$ represent aerodynamic drag force, rolling resistance, road grade force, and traction force, respectively. The acceleration force is expressed as follows:

$$F_{acc} = ma \tag{2}$$

where $m$ is vehicle weight and $a$ is vehicle acceleration. The vehicle acceleration is correlated to the vehicle speed:

$$a = \dot{v}_{veh} \tag{3}$$

The rest of the four forces are calculated as follows:

$$\begin{cases} F_{air} = \frac{1}{2}\rho C_d A v_{veh}^2 \\ F_{roll} = cos(\beta)\, f_{roll} mg \\ F_{grade} = sin(\beta)\, mg \\ F_{trac} = F_{ICE} + F_{EM} \end{cases} \tag{4}$$

where $\rho$ is the air density, $C_d$ is the air drag coefficient, $A$ is vehicle font area, $v_{veh}$ is vehicle speed, $\beta$ is the road slop angle, $f_{roll}$ is road rolling resistance coefficient, $m$ is the weight of the vehicle, $g$ is the gravity constant, $F_{ICE}$ and $F_{EM}$ are the traction force provided by engine and EM, respectively.

The engine force is derived as follows:

$$F_{ICE} = \frac{T_{whl,front}}{r_{whl}} \tag{5}$$

$$T_{whl,front} = T_{trans,in} r_{fd} r_{gear} \eta_{trans} \tag{6}$$

$$T_{trans,in} = T_{ICE} \tag{7}$$

Engine fuel consumption is interpolated from the fuel map (Fig. 2 (a)) as follows:

$$\dot{m}_{ICE} = f(\omega_{ICE}, T_{ICE}) \tag{8}$$

$$\omega_{ICE} = \omega_{trans,in} \tag{9}$$

$$\omega_{trans,in} = \omega_{whl,front} r_{fd} r_{gear} \tag{10}$$

$$\omega_{whl,front} = \frac{v_{veh}}{r_{whl}} \tag{11}$$

where $r_{whl}$ is wheel radius and it applies for front and rear wheels, $r_{fd}$ is the gear ratio of the final drive, and $r_{gear}$ is the gear ratio of transmission. The EM force is calculated as follows:

$$F_{EM} = \frac{T_{whl,rear}}{r_{whl}} \tag{12}$$

$$T_{whl,rear} = T_{EM} r_{EM} \eta_{EM,mech} \tag{13}$$

Battery power output calculation is as follows:

$$P_{bat} = \begin{cases} \frac{\omega_{EM} T_{EM}}{\eta_{EM,elec}}, discharge \\ \omega_{EM} T_{EM} \eta_{EM,elec}, charge \end{cases} \quad (14)$$

$$\eta_{EM,elec} = f(\omega_{EM}, T_{EM}) \quad (15)$$

$$\omega_{EM} = \omega_{whl,rear} r_{EM} \quad (16)$$

$$\omega_{whl,rear} = \frac{v_{veh}}{r_{whl}} \quad (17)$$

where the EM efficiency map is shown in Fig. 2(b). Total vehicle torque demand is determined by the driver command (acceleration pedal position) as follows:

$$T_{dmd} = \theta_{acc} T_{ref} \quad (18)$$

The total torque demand is split by the energy management system, which is Q-learning in this study as follows:

$$T_{dmd} = T_{dmd,ICE} + T_{dmd,EM} \quad (19)$$

The total EM torque demand includes the torque demand assigned by supervisory control and braking torque demand:

$$T_{brake} = \theta_{brake} T_{EM,min} \quad (20)$$

$$T_{dmd,EM,total} = T_{dmd,EM} + T_{brake} \quad (21)$$

The driver model is a PID feedback control, which takes actual vehicle speed and reference speed as the inputs, and outputs a bounded value within the range of [-1,1]. The brake pedal position is activated when the PID output is negative, and the acceleration pedal position is activated when PID output is positive. -1 and 1 are referred to the full positions of the brake pedal and acceleration pedal, respectively.

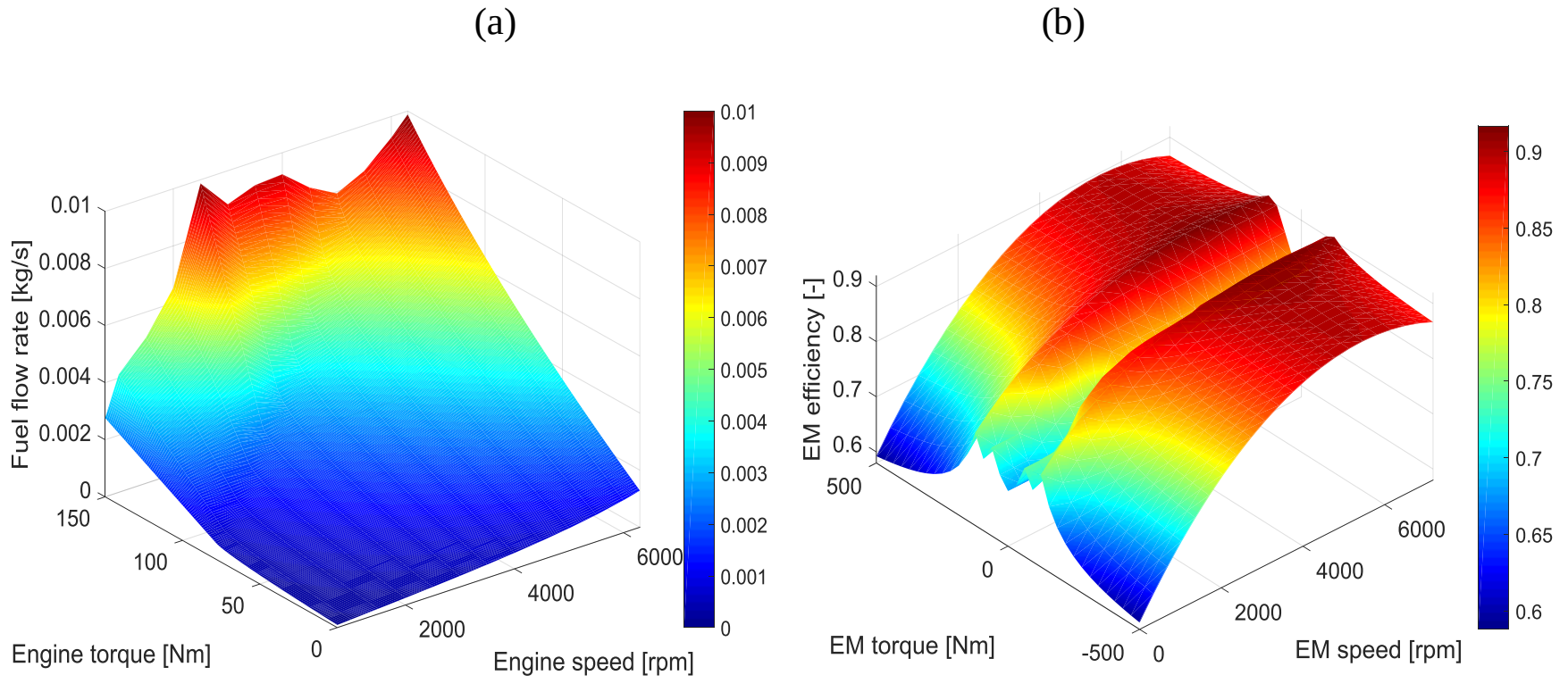

Fig. 2. (a) Engine fuel flow rate map and (b) EM efficiency map.

## B. Battery Modeling

The time derivative of the battery state of charge (SOC) is calculated as follows:

$$S\dot{O}C = -\frac{I_{bat}}{Q_{bat}} \quad (22)$$

where $I_{bat}$, $Q_{bat}$ are the battery current and nominal capacity, respectively. Current and voltage are calculated as follows:

$$I_{bat} = \frac{P_{bat}}{U_{bat}} \tag{23}$$

$$U_{bat} = U_{oc} - I_{bat}R_{internal} \tag{24}$$

where $P_{bat}$ is the battery power output, $U_{oc}$ is the battery open circuit voltage, and $R_{internal}$ is the battery internal resistance.

## III. SUPERVISORY CONTROL STRATEGIES

### A. Q-learning strategy

In the RL framework, there are two objects: the environment on the left and agent on the right in Fig. 3. The agent is the RL Q-learning algorithm. Everything outside the agent belongs to the environment. In this application, the vehicle is the main object in the environment. Besides the vehicle, any factor influencing vehicle operating status is also integrated into the environment block, such as driver behavior and road conditions. The agent aims to maximize the reward by learning from the interaction with the environment through the data flow path. The interaction contains the action to the environment and feedback from the environment.

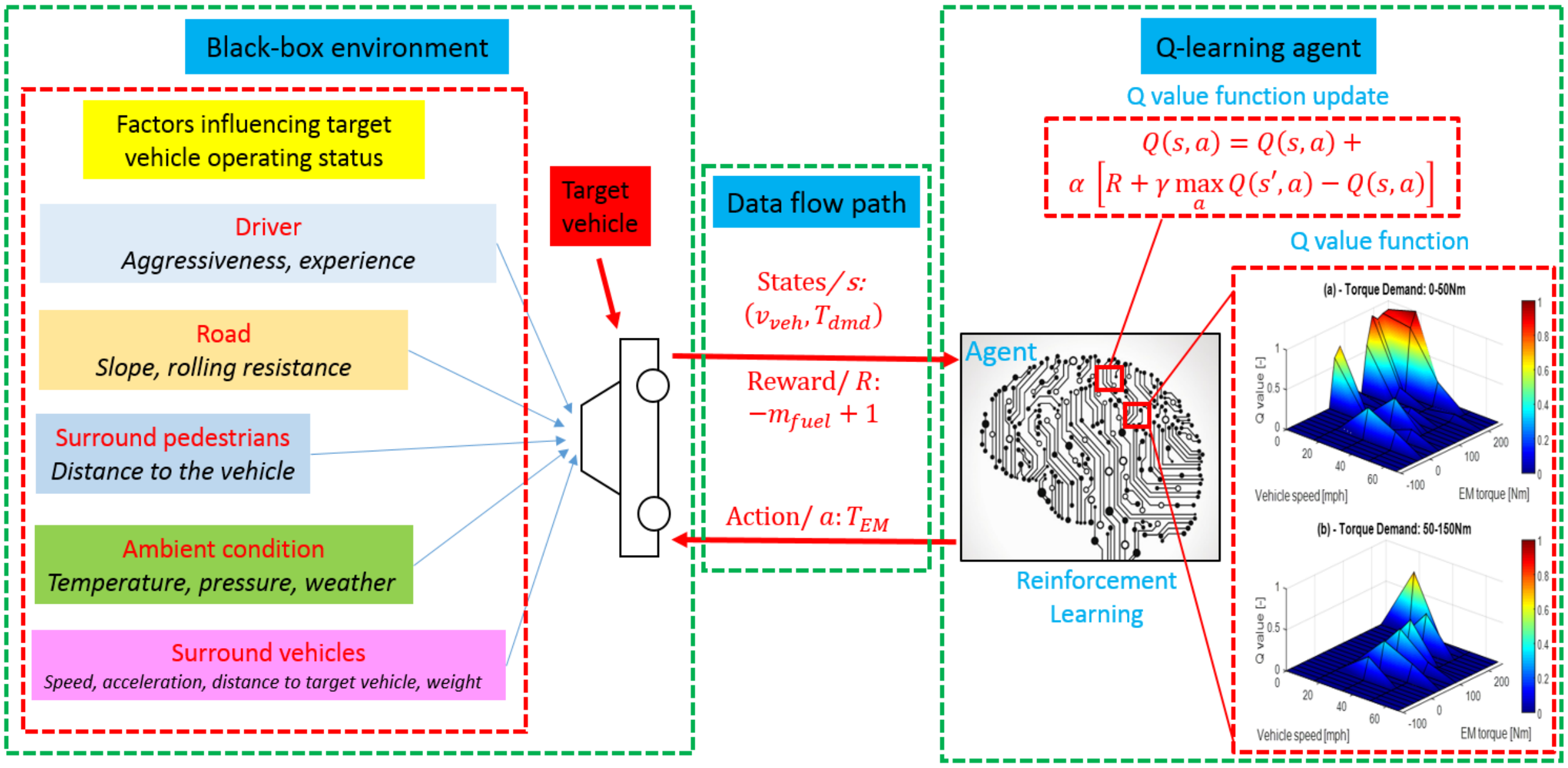

Fig. 3. Q-learning EMS and vehicle interaction diagram.

The Q-learning algorithm learns the state-action value function – $Q(s,a)$. The $Q(s,a)$ is a function of state $s$ and action $a$. Its value represents the overall reward the agent can get if the respective action is taken at the given state. The learning process is basically improving the Q value estimation accuracy by correcting the old estimated Q value with the observed error (i.e. representing Dopamine in a human brain). The Q value correction process is as follows:

$$Q(s,a) = Q(s,a) + \alpha\left(\left[R + \gamma \max_a Q(s',a)\right] - Q(s,a)\right) \tag{25}$$

where $\alpha$ is the learning rate, $s$ is the state from current time step, $s'$ is the state from next time step, $R$ is the reward obtained in state transition from $s$ to $s'$, $\gamma$ is the discount factor. The term $\left[R + \gamma \max_a Q(s', a)\right]$ represents the observed overall reward respective to $(s, a)$ state-action pair, and it is from Bellman Equation [9]. To be specific, the reward $R$ represents the reward from state transition from $s$ to $s'$ and $\max_a Q(s', a)$ represents the max overall reward from $s'$ to the infinite future horizon. Adding these two terms together represents the overall reward from $s$ to the infinite future. The discount factor $\gamma$ aims to reduce the weight of future term $\max_a Q(s', a)$ as there could be uncertainty in the future or instantaneous reward $R$ could be what the agent cares.

Table 2 shows the pseudocode of the Q-learning algorithm used in this study. There are four main processes: (i) experience exploration, (ii) experience evaluation, (iii) Q value update, and (iv) experience evaluation criteria update. Experience exploration aims to try some random actions and generate new experiences so that the Q-learning performance gets improved. If the warm start method is used, the first iteration of experience exploration (i.e., i=1) is executed using warm start supervisory control. If ECMS is used, ECMS directly acts as supervisory control in the first iteration. If heuristic supervisory control is used, the Q values are modified with the heuristic rules, and then the modified Q values are used in supervisory control using $\varepsilon$-greedy method with $\varepsilon = 0$. Experience evaluation aims to evaluate the quality of the experience and filter the experience used for Q-learning state-action value function update. Q value update aims to update the Q values using the selected experience. Experience evaluation criteria update aims to choose the best criteria based on the latest knowledge so that the experience selection process gradually gets improved. More information about experience evaluation can be found in [30]. In this vehicle application, one iteration is one simulation over an entire driving cycle. $\varepsilon$-greedy method is used in the random action exploration, and $\varepsilon$ is in the range of [0,1]. More specifically, in the vehicle operation, random action is selected with the possibility of $\varepsilon$ and the optimal action (i.e., the action corresponding to the largest Q value among all the Q values in the given state) is selected with the possibility of (1- $\varepsilon$).

TABLE 2.

Pseudocode of Q-learning algorithm implemented in this study.

**Q-learning Algorithm**

Initialize $Q(s, a)$ with zeros, for all $s \in S$, $a \in A(s)$.

  Initialize $R_{tot}$ with zero.

**for** $i \in (1, \dots, N)$ **do** (for each iteration):

  **Experience exploration:**

  **if** (warm start==1 & i==1) **do**

    Initialize s

    **for** $j \in (1, \dots, M)$ **do** (for each time step of iteration):

      Take action $a_j$ using warm start supervisory controls, observe $R_j$, $s_{j+1}$ from environment (i.e., vehicle model)

    **end for**

**else**
Initialize s
**for** $j \in (1, \dots, M)$ **do** (for each time step of iteration):
Choose action $a_j$ at state $s$ using policy derived from $Q$
($\varepsilon$-greedy action selection method, $\varepsilon = 0.1$)
Take action $a_j$, observe $R_j$, $s_{j+1}$ from environment
**end for**
**end if**
**Experience evaluation:**
**if** $\sum_1^M R_j > R_{tot}$ **do**
**Q value update:**
**for** $j \in (1, \dots, M)$ **do** (for each time step of iteration):
$Q(s_j, a_j) = Q(s_j, a_j) + \alpha \left[ R_j + \gamma \max_a Q(s_{j+1}, a) - Q(s_j, a_j) \right]$
**end for**
**Experience evaluation criteria update:**
Initialize s
**for** $j \in (1, \dots, M)$ **do** (for each time step of iteration):
Choose action $a_j$ at state $s$ using policy derived from
$Q$ ($\varepsilon$-greedy action selection method, $\varepsilon = 0$)
Take action $a_j$, observe $R_j$, $s_{j+1}$ from environment
**end for**
$R_{tot} = \sum_1^M R_j$
**end if**
**end for**

---

### *1) Reward*

In RL, the reward is the metric of how good the action is, and it is the result of state transition from the current state to the next state. In Q-learning, the reward combined with an estimated future return is treated as the observed overall return, which is compared with the estimated overall return. The error from the comparison is considered to mimic the Dopamine during the new task learning in the human brain [31]. In this study, the fuel consumption is defined as the reward. The goal of the supervisory control is to maximize the reward.  Battery usage is represented by the equivalent fuel consumption as follows:

$$\dot{m}_{fuel,bat} = \left( cS_{dis} \frac{1}{\eta_{dis}} + (1-c) S_{chg} \eta_{chg} \right) \frac{P_{bat}}{\mathrm{Q_{LHV}}} \tag{26}$$

$$c = \begin{cases} 1 & , discharge \\ 0 & , charge \end{cases} \tag{27}$$

where $S_{dis}$ and $S_{chg}$ represent the discharge and charge equivalent factors, $\eta_{dis}$ and $\eta_{chg}$ represent the discharge and charge efficiency, $P_{bat}$ is the power output/ input of the battery, and $Q_{LHV}$ is the low heating value of the fuel used by engine. The discharge and charge equivalent factors are assumed to be 3, and efficiencies are assumed to be 0.95. The reward expression is as follows:

$$R = -(\dot{m}_{fuel,ICE} + \dot{m}_{fuel,bat})dt + 1 \tag{28}$$

where $dt$ is the time step in the simulation. A negative sign of fuel consumption turns the fuel minimization problem into a maximization problem, as the Q-learning is implemented in reward maximization problems by default. The addition of 1 in (28) ensures the reward to be a positive value in any conditions. 1 is enough in this vehicle as the first term of right-hand side in (28) is in the magnitude of 0.001. In other large vehicle applications, a great number can be used, such as 10 or 100. The positive reward attracts the greedy action selection at the beginning phase of the learning as the initial Q values are zeros. Otherwise, if the reward is a negative value, the greedy policy will choose non-visited Q values over visited negative Q values, which diverts the HEV from ideal performance in the learning process.

#### *2) States*

States in RL represent the status of the environment and is one of the two feedback signals from the environment (the other signal is the reward). The RL action heavily depends on states. In this study, vehicle torque demand and vehicle speed are selected as the state vector. Vehicle torque demand is related to the vehicle traction force and acceleration. The vehicle torque demand selection over the vehicle acceleration is due to the direct connection between torque demand and action (i.e., EM torque demand) without unit change. Vehicle speed represents the accumulative vehicle acceleration. Different state vector examples in HEV RL-based supervisory controls can be found in literature [11, 12, 16, 17]. State vector selection should be systematically conducted and is not the focus of this study. Detailed analysis can be found in [30].

#### *3) Action*

The EM torque demand is chosen as the action in this study. The engine torque demand is then derived as the difference between the vehicle torque demand and EM torque demand. In order to maximize the vehicle efficiency, regenerative braking is active for all the braking events. The rest of the actions are done by local controllers, such as transmission shift and torque converter lock. $\varepsilon$-greedy method is used in the random action exploration and $\varepsilon$ is in the range of [0,1].

### *B. Warm start strategies*

#### *1) Heuristic strategy*

The engine generally has low efficiency when the vehicle has a large acceleration at vehicle start, which frequently occurs in urban driving scenarios. During this period, vehicle torque demand is great, and the torque demand magnitude varies with vehicle classes. In the vehicle of this study, the engine saturates at maximum torque during vehicle start, and the engine efficiency is not optimal. More specifically, the engine efficiency map utilized in this study is shown in Fig. 4. When engine speed is less than 2800rpm, the maximum engine efficiency at the torque saturation line is 28%. However, the maximum engine efficiency reaches 32% on the entire map. The idea of the heuristic strategy is to split part of the torque demand to EM during the vehicle fast acceleration at low speed so that the engine could operate at slightly lower torque and thus higher efficiency. To implement this heuristic strategy, the Q values are modified. First, the Q values are initialized to be zeros at all states and actions. Then, the Q values at low speed, high torque demand, and large EM torque are set to be 1.0, which guides the Q-learning to select large EM torque in this scenario.

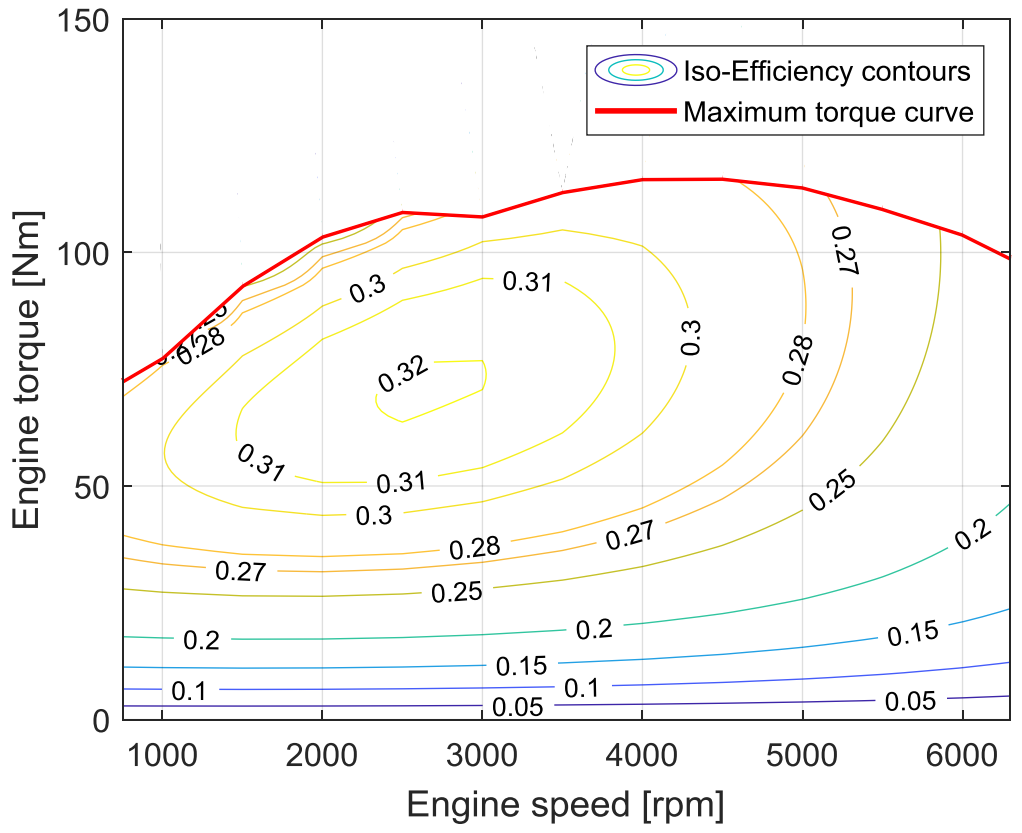

Fig. 4. Engine efficiency map.

### 2) *Equivalent Consumption Minimization Strategy*

Equivalent Consumption Minimization Strategy (ECMS), as one of the well-known supervisory controls, considers both engine fuel consumption and battery energy usage and minimizes the overall equivalent fuel consumption. It uses (26) to calculate the equivalent fuel consumption in a battery. In real-time vehicle operation, ECMS optimizes the torque split between the engine and EM at the given speed and driver torque demand based on the efficiency maps of engine/ EM.

## IV. SIMULATION AND RESULTS ANALYSIS

Urban Dynamometer Driving Schedule (UDDS) driving cycle [32] is utilized in the learning as the vehicle operating conditions. The simulation time step is set to be 1 second. The total number of iterations is set to be 5000. The Q value update only occurs when the sum of rewards from iteration is greater than that from simulation applying full greedy action. Full greedy action simulation is conducted after Q values are updated, the result of which is used as criteria to select the good experience for Q value update. The learning rate $\alpha$ is fixed at 0.1, and the random action exploration rate $\varepsilon$ is fixed at 0.1 (i.e. 10%). The discount factor $\gamma$ is set to 1.0. For the discretization of states, vehicle torque demand is discretized in 5 values from 0 to 400Nm in 100Nm steps.

Vehicle speed is discretized in 5 values from 0 to 65mph in 16mph steps. For action discretization, EM torque is discretized in 20 values from -100 to 250Nm in 18.5Nm steps.

The results from cold start Q-learning are presented first as the reference to the warm start methods. Then the results from the warm start using ECMS are analyzed and compared with cold start results. After that, a heuristic rule of EMS is converted to Q values initialization as another warm start method, and the results are compared with the cold start and ECMS warm start.

## 4.1 *Learning with cold start*

In the cold start Q-learning, the Q values are initialized with zeros. No knowledge is utilized to help initialize the Q values. The learning process is executed in 5000 iterations. Only the iterations generating good experience are utilized to update the Q values. The theoretical maximum (calculated in experience evaluation criteria update in Table 2) of the sum of rewards and engine fuel economy along the 5000 iterations are shown in Fig. 5. From the figure, it is observed that the learning converges after 600 iterations. In Fig. 5(b), the fuel economy starts with 37mpg and converges to 59mpg, which presents a 59.5% improvement.

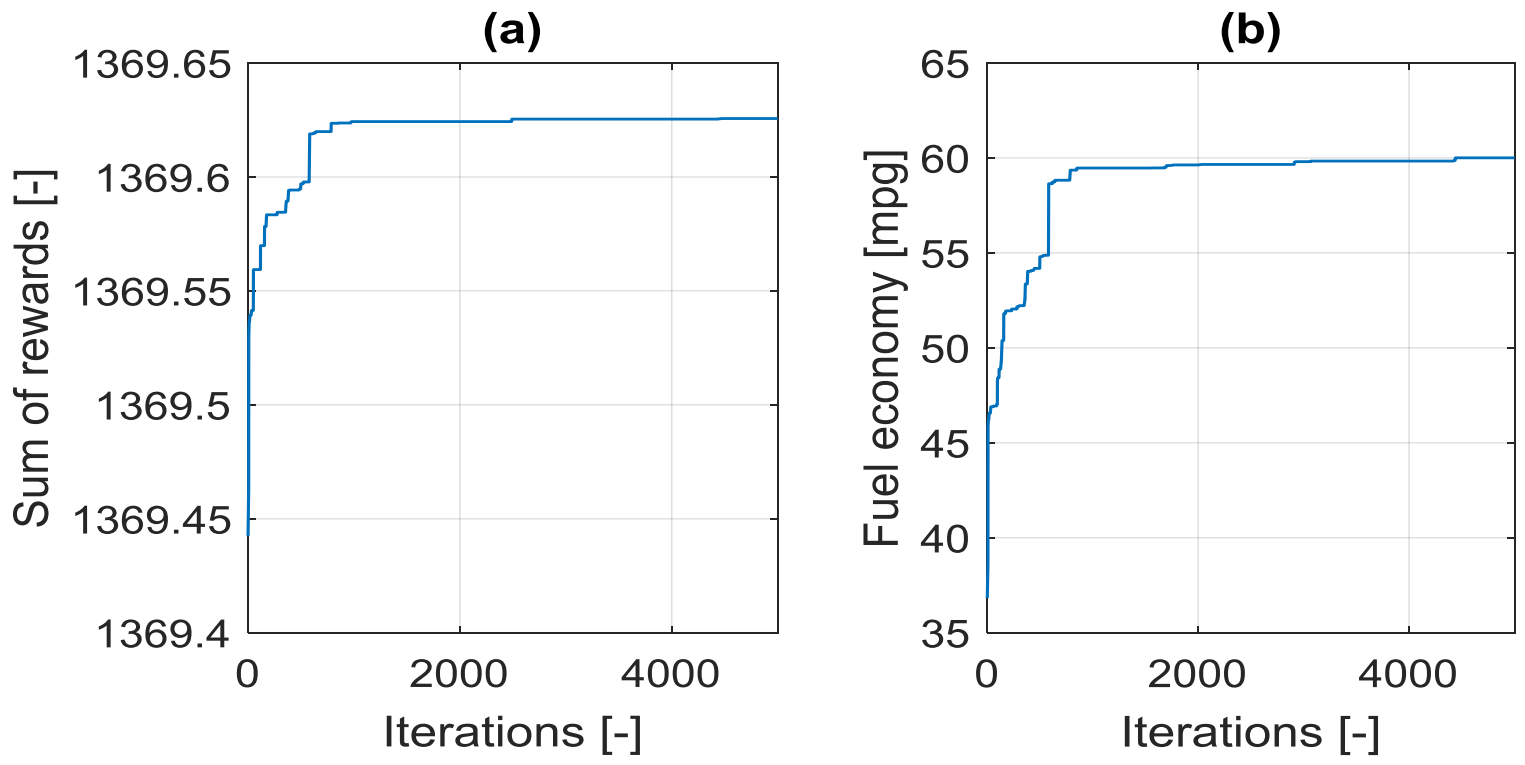

Fig. 5. Sum of rewards and fuel economy from good experience along the 5000 iterations of cold start Q-learning.

## 4.2 *Learning with warm start*

In the ECMS warm start, the vehicle EM and engine torque split are controlled by ECMS in the first UDDS cycle. After this one cycle simulation. The result is used as an experience to update the Q values following the process in Table 2. Prior to the update, the Q values are all zeros. After the update, the Q values are the ECMS warm start results, which then undergo 5000 iterations.

Different from ECMS warm start, the heuristic warm start does not require the single UDDS simulation. First, all the Q values are initialized to be zeros. Then, the Q values at low speed and high torque demand are set to 1.0, which guides the Q-learning to select large EM torque in this scenario. After the warm start, the Q values then undergo 5000 iterations.

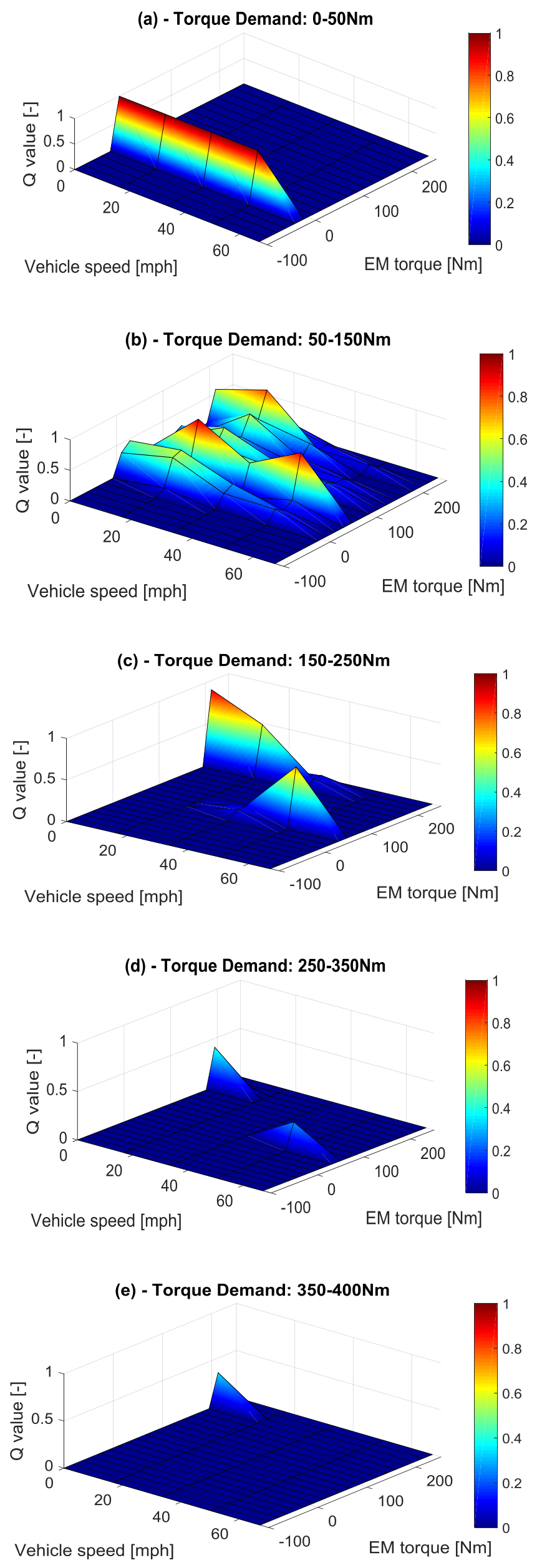

Fig. 6. ECMS warm start Q values initialization. Q value is as a function of speed and torque at different torque demand.

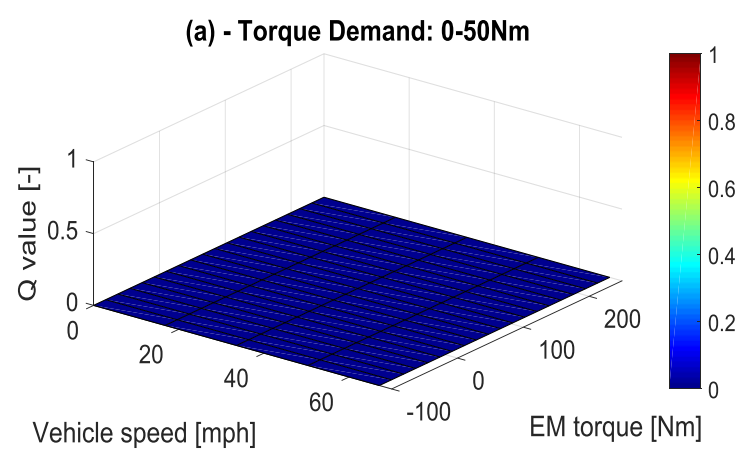

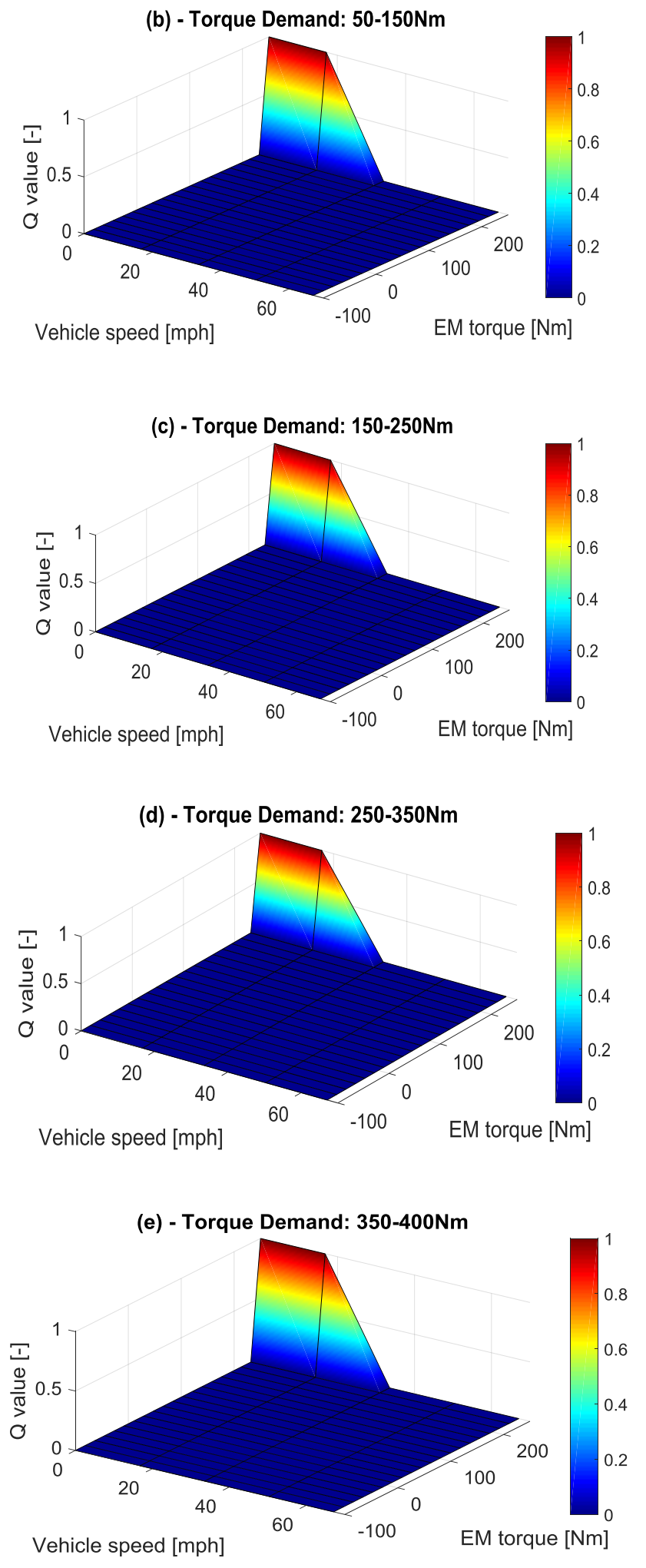

Fig. 7. Heuristic warm start Q values initialization. Q value is a function of speed and torque at different torque demand.

Before the iterations, the initialized Q values from ECMS and heuristic warm start strategies are shown in Fig. 6 and Fig. 7, respectively. In the ECMS warm start Q values, main updates occur in the 50-150Nm torque demand range, as shown in Fig. 6(b). In this subplot, there are multiple Q value updates (i.e., multiple EM torques action taken) at each vehicle speed. In Fig. 6(a), when total torque demand is below 50Nm, EM torque from ECMS is close to zero along with all the vehicle speed. At the top of Fig. 6(c), (d) and (e), there are Q value spikes, which are corresponding to large EM torque at low speed (<20mph) and high torque demand (150-400Nm). This scenario with low speed and high torque demand means the large acceleration at low speed.

For the heuristic warm start, the idea is similar to the observation from Fig. 6(c), (d), and (e), where EM helps vehicle acceleration at low speed and large acceleration scenario. There are two reasons to use EM in this scenario: (i) engine maximum torque is not large enough to satisfy the torque demand, (ii) engine efficiency at engine torque saturation line is not high. Thus the EM could help improve the overall efficiency of the propulsion system. When the vehicle speed is less than 20mph and torque demand is greater than 50Nm, Q value is assigned with 1.0, as shown in Fig. 7(b-e). By integrating all the subplots in Fig. 6 by selecting the maximum Q value at each state, the optimal action map is obtained in Fig. 8(a). Similarly, the optimal action map from heuristic initialization is obtained and shown in Fig. 9(a). The main trend of these two optimal action maps is the same that the high EM torque is used during low vehicle speed while low EM torque is used during high vehicle speed. The optimal EM torque in Fig. 8(a) is slightly different at different total torque demand, while this relationship is not observed in Fig. 9(a). In addition, some middle-range EM torques appear in the middle to high speed regions in Fig. 8(a), which reduces the burden of the engine. While the EM torque is zero in those speed range in Fig. 9(a). These differences result in the ECMS and heuristic initialization fuel economy difference (51mpg vs. 42mpg).

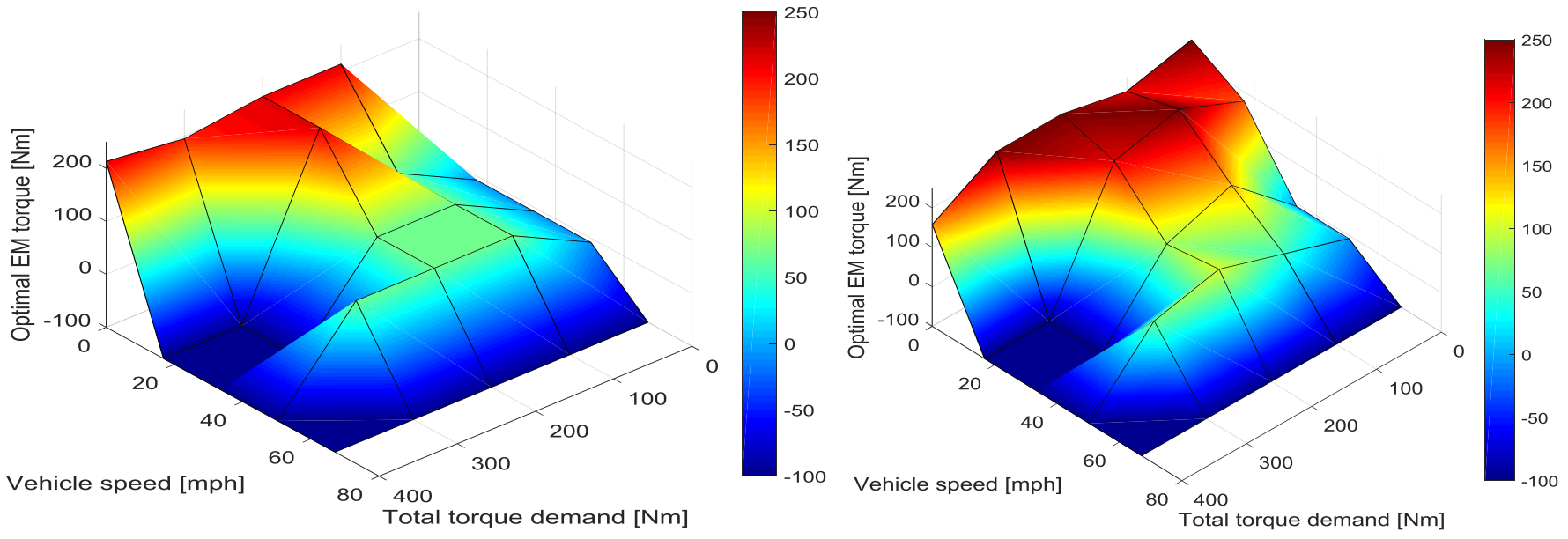

Fig. 8. Optimal EM torque comparison between initial Q values and end Q values of ECMS warm start iterations: (a) initial Q values of warm start, and (b) end Q values of warm start after 5000 iterations.

(a) Heuristic Initialization (b) Heuristic after 5000 Iterations

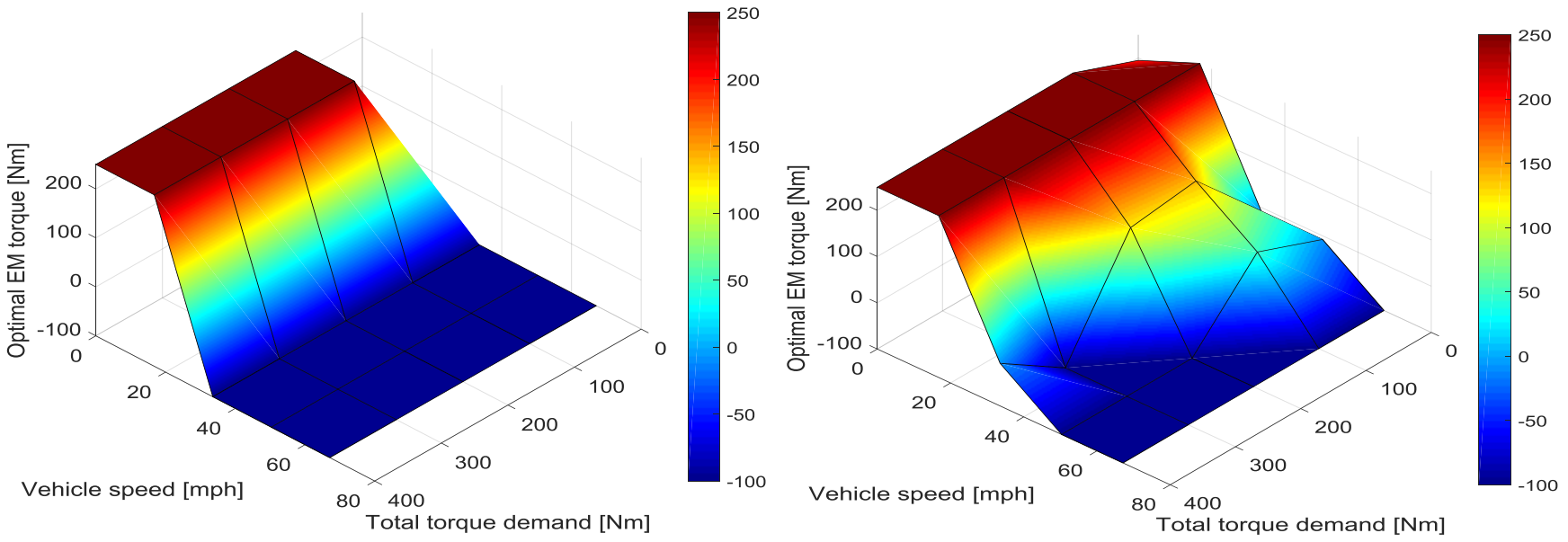

Fig. 9. Optimal EM torque comparison between initial Q values and end Q values of heuristic warm start iterations: (a) initial Q values of warm start, and (b) end Q values of warm start after 5000 iterations.

The operational efficiencies of engine and EM using cold start, ECMS warm start, and heuristic warm start Q-learning are shown in Fig. 10, Fig. 11 and Fig. 12, respectively. When compared with Fig. 10(a), Fig. 11(a) has less operating points on the saturation torque line (i.e., low-efficiency region) and more operating points in the highlighted high-efficiency region. The high efficiency is beneficial to the fuel economy as the ECMS warm start Q-learning results in 51mpg while the cold start Q-learning results in 36mpg. This improvement significantly improves the overall fuel economy during the learning process. The high torque demand from the ICE in Fig. 10(a) is satisfied by the EM in the highlighted high torque region of Fig. 11(b). The engine operating points disparity between Fig. 10(a) and Fig. 12(a) is not obvious. The disparity is more obvious when comparing the EM operating torque between Fig. 10(b) and Fig. 12(b) as more large positive torque operating points locate in the highlighted region of Fig. 12(b). The high EM torque is the main principle of the heuristic warm start strategy. The ECMS and heuristic warm start Q-learning show similar EM operating efficiency. The main benefit of large EM torque is to reduce engine torque saturation and thus increase engine operating efficiency.

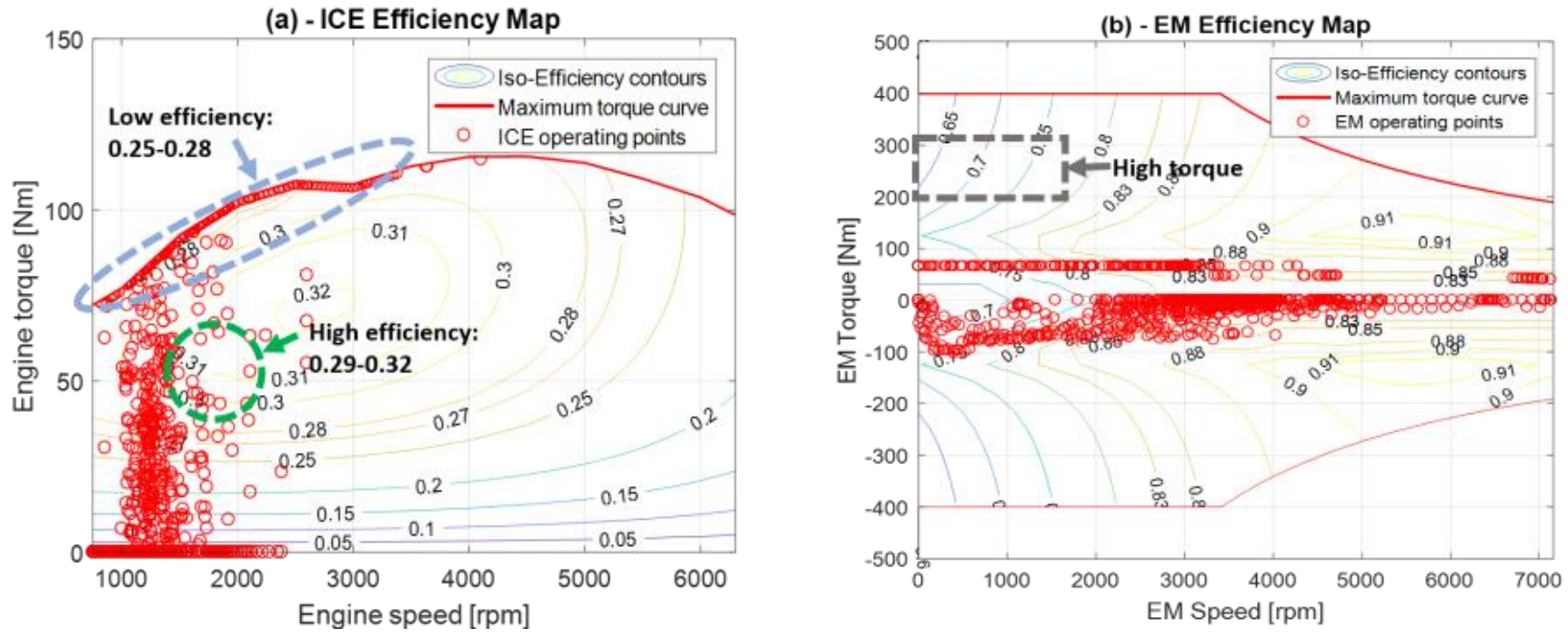

Fig. 10. ICE and EM operating efficiency from the UDDS driving cycle using all-zero Q values (cold start) prior to 5000 iterations.

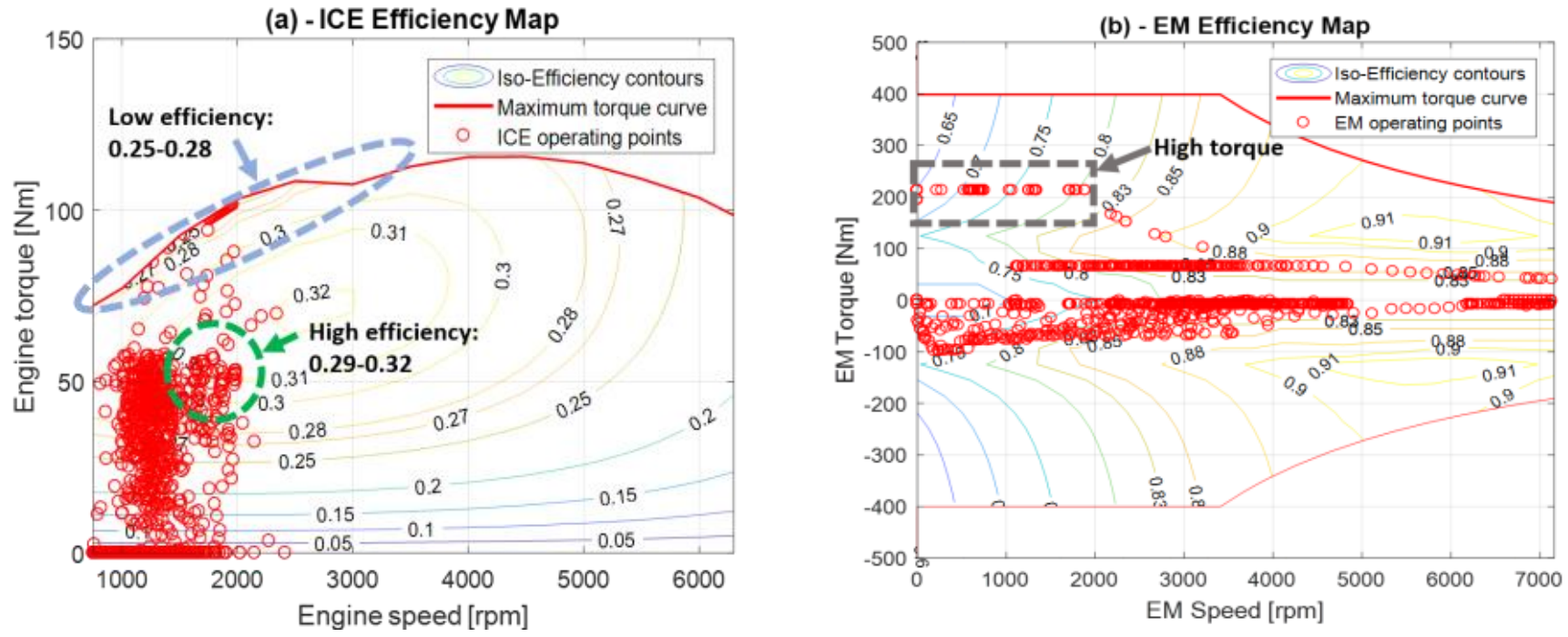

Fig. 11. ICE and EM operating efficiency from the UDDS driving cycle using ECMS initialized Q values (warm start) prior to 5000 iterations.

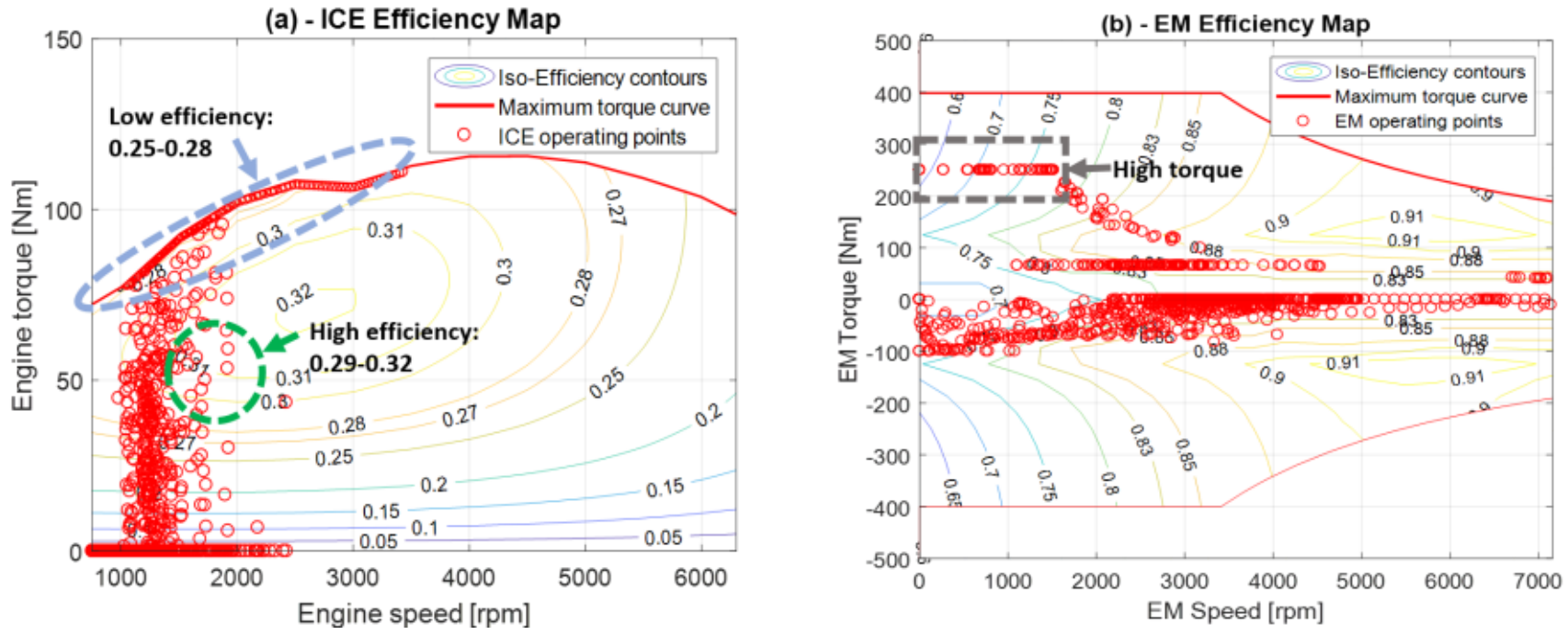

Fig. 12. ICE and EM operating efficiency from the UDDS driving cycle using heuristic initialized Q values (warm start) prior to 5000 iterations.

After the warm start initialization, 5000 iterations are conducted. The sum of rewards comparison between cold start and warm start from good experience along the iterations are shown in Fig. 13. Both warm start strategies give Q-learning a tremendous leading position at the initial iteration of 5000 iterations when compared with cold start. The fuel economy comparisons of cold start and warm start are shown in Table 3. The fuel economies at iteration 1 from ECMS and heuristic warm start are 51mpg and 42mpg, respectively, while the cold start only produces 36mpg. It takes 800 iterations for the cold start Q-learning to reach 59mpg, while the ECMS warm start Q-learning only requires 250 iterations to reach 59mpg, which is 68.8% iterations reduction. The distance of UDDS driving cycle is around 7.5miles, and the duration is 1369s. If the car OEMs design the EMS based on Q-learning, for the cold start, it requires 6000 miles/ 12.5 days of test driving to achieve 59mpg, while warm start only needs 1875 miles/ 4 days of test driving, which is a significant reduction in test time and cost. For a car owner to train the Q-learning based on the 60 miles daily commute distance, it takes 3.5 months to achieve 59mpg using cold start initialization strategy, while only takes only one month to reach a similar fuel economy using ECMS warm start. For the heuristic warm start, it takes a similar amount of iterations with ECMS warm start to achieve 59mpg.

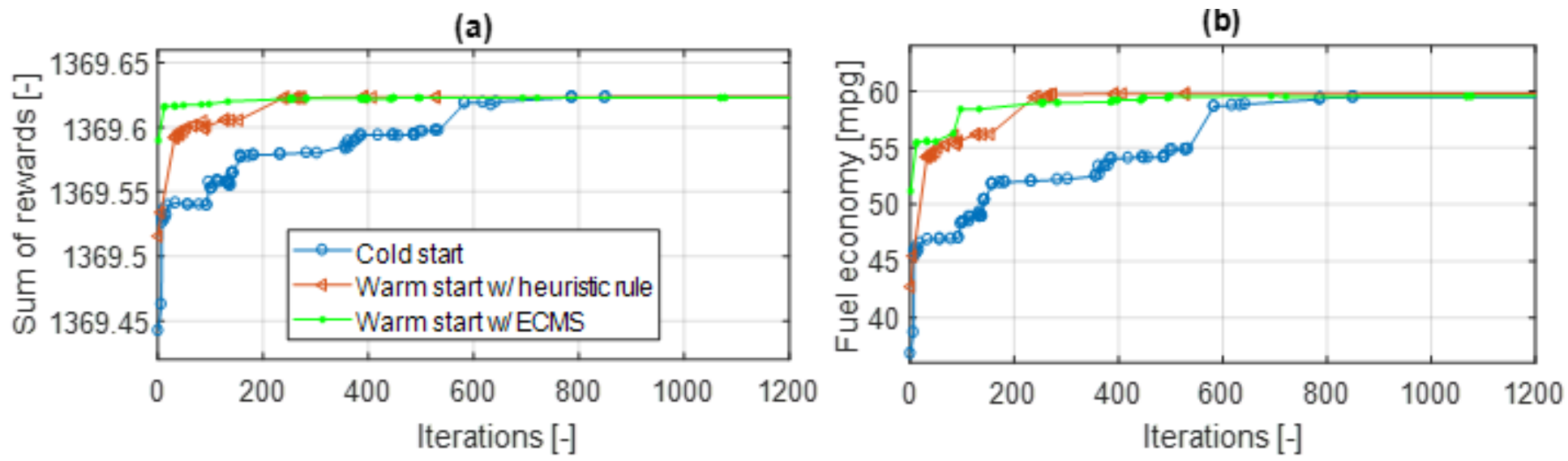

Fig. 13. Comparison of good experience obtained along the iterations from cold start and warm start: (a) sum of rewards, and (b) fuel economy. (mean of 5 runs)

Table 3.

Fuel economy of ECMS and heuristic warm start initialization and final results of Q-learning in UDDS driving cycle.

| Learning stage | Cold start Q-learning | Heuristic warm start Q-learning | ECMS warm start Q-learning |
|---|---|---|---|
| Initialization (1st iteration) | 36.6 MPG | 42 MPG | 51 MPG |
| Final (5000th iteration) | 59.4 MPG | 59.6 MPG | 59.5 MPG |

After 5000 iterations, the Q value maps from ECMS warm start and heuristic warm start are shown in Fig. 14 and Fig. 15, respectively. Both Q value maps share the shapes in each of the total torque demand subplots. The similarity of Q value explains the similar fuel economy performance of the two Q value initialization strategies. When comparing the initialized Q values (Fig. 6/ Fig. 7) and the Q values after 5000 iterations (Fig. 14/ Fig. 15), the shape from the initialization is still visible with significantly dampened magnitude. For instance, the spike at 0 EM torque in Fig. 6(a) still exists at the same location of the map in Fig. 14(a). However, the Q value decreases from 1 to 0.1. A similar observation can be found in Fig. 7(c) and Fig. 15(b). The reason for decreasing magnitude is the multiple updates during the 5000 iterations. Every time the maximum Q value in the entire Q value table is increased, the magnitude of initialized Q value is discounted relative to the maximum Q value. Combining the subplots of Q values by only selecting the maximum Q value along with actions, optimal action maps from ECMS and heuristic warm starts are obtained in Fig. 8(b) and Fig. 9(b), respectively. Comparing Fig. 8(a) and (b), one can easily tell the reason why ECMS warm start initialization has 51mpg, which is its similarity shape of optimal action map to the final map after 5000 iterations. In comparison, heuristic warm start optimal action maps show more differences between Fig. 9(a) and (b), which is corresponding to the 17 mpg difference (42mpg vs. 59mpg).

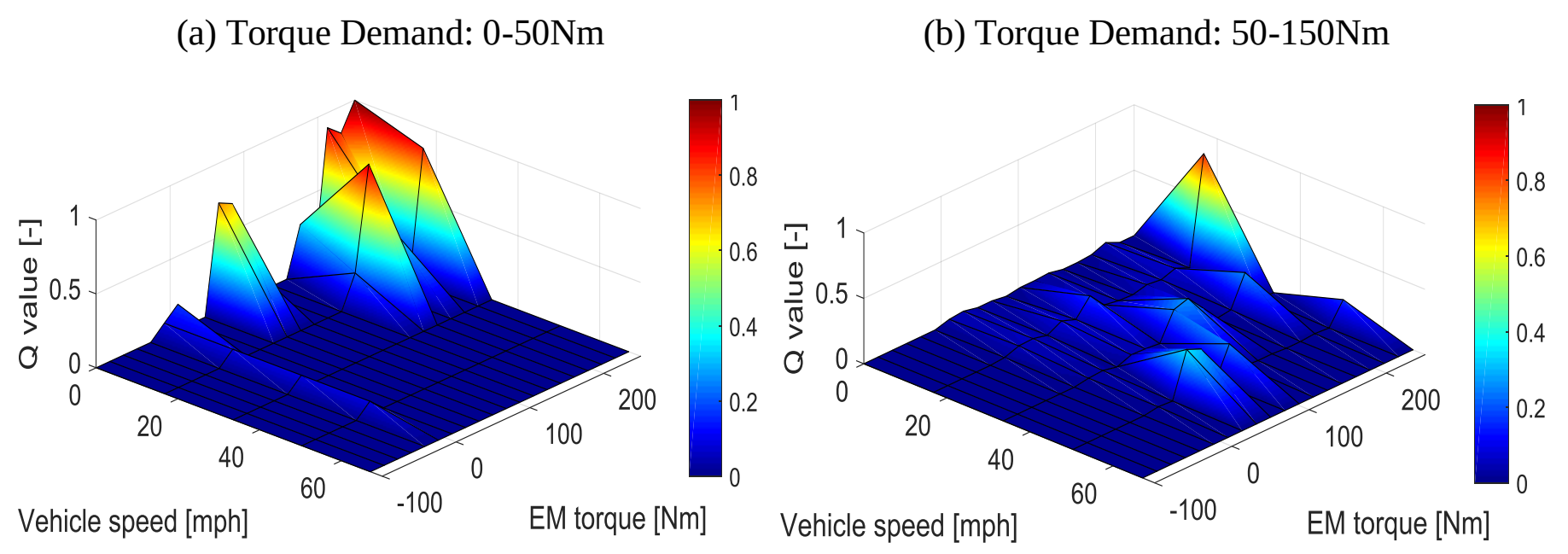

Fig. 14. ECMS warm start Q values after 5000 iterations. Q value is as a function of speed and torque at different torque demand.

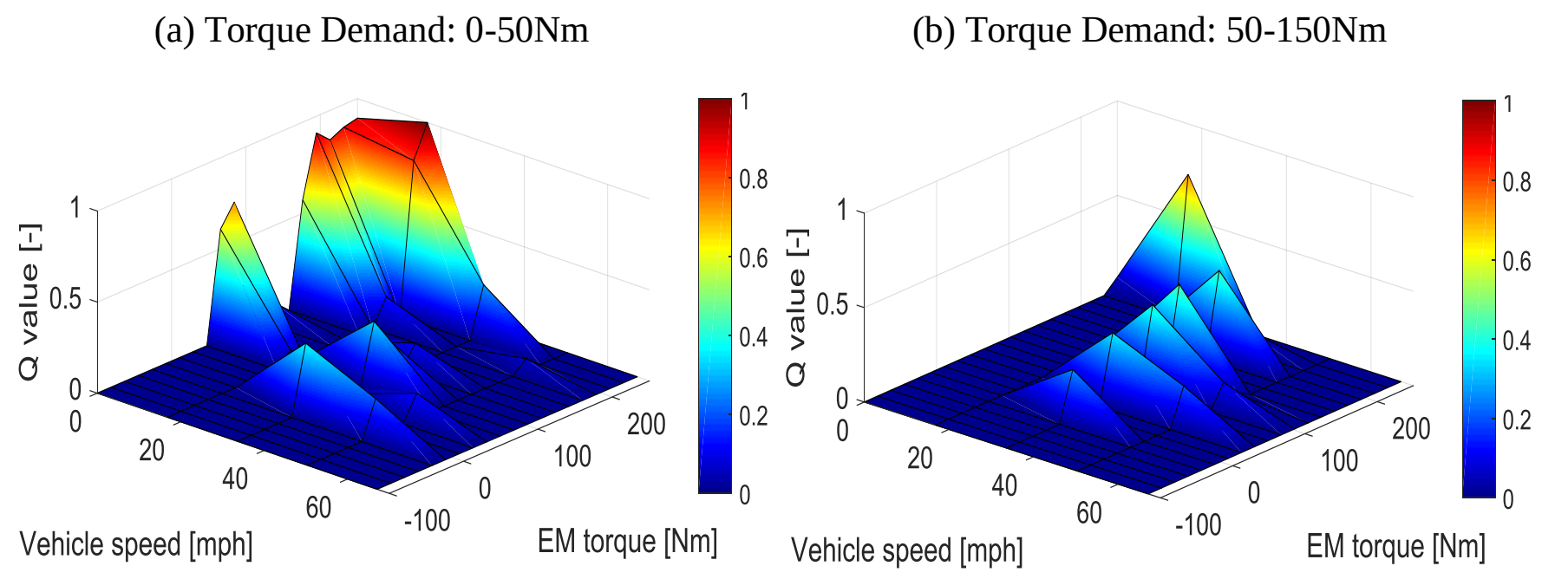

Fig. 15. Heuristic warm start Q values after 5000 iterations. Q value is a function of speed and torque at different torque demand.

(a) ECMS Initialization (b) ECMS after 5000 iterations

Simulation results comparison among cold start initialization (iteration #1), ECMS warm start initialization (iteration #1), heuristic warm start initialization (iteration #1) and heuristic warm start after 5000 iterations (iteration #5000) are shown in Fig. 16 and Fig. 17. In Fig. 16(a), all four simulations show satisfactory vehicle speed tracking performance. Battery SOC results are shown in Fig. 16(b), where cold and heuristic initialization simulations have similar SOC trajectories. ECMS warm start initialization and heuristic warm start final show large SOC operating range. The starting SOC is 0.6, and the ending SOC for all three cases is between 0.58-0.6. Fig. 16(c-d) show the engine fuel rate and cumulative fuel consumption along the driving cycle. In Fig. 16(c), multiple peak fuel rates come from the cold start initialization, which explains its high cumulative fuel consumption through the entire driving cycle. ECMS and heuristic warm start initialization burn less fuel than cold start. Torque information is considered for comparison in Fig. 17, where a zoom-in time window (400s-500s) result is plotted for better readability. In this 100s time window, the vehicle starts and stops twice, as shown in Fig. 17(a). There are two moments (410s and 460s) where the cumulative fuel consumption gaps increase among the four simulation cases, as shown in Fig. 17(d). The largest fuel consumption increase occurs in cold start case. At both moments, the vehicle starts to accelerate. In the first 1-2 seconds of the vehicle start at 401s, the total torque demand for all four cases is similar, as shown in Fig. 17(e). At this vehicle start scenario, all four cases split a similar amount of torque to the engine, while the EM torques are different, as shown in Fig. 17(f) and (g). For the EM torque, the cold start only assigns 65Nm, which is the default action when the Q values are the same for the entire 20 action discretizations. However, the ECMS warm start and heuristic warm start specifically initialize the Q values based on the low speed and high total torque demand scenario so that EM can boost the vehicle during vehicle start. Thus, ECMS warm start initialization and heuristic warm start initialization, assign 200-250Nm (high torque) during the vehicle start. The high EM torque reduces the burden of the engine torque output, as shown in Fig. 17(g) and thus reduce engine fuel rate. The low EM torque in the cold start results in a slightly large speed tracking error, as shown in Fig. 17(a), and thus leads to higher total torque demand in Fig. 17(e) as the "PI" controller driver tries to add more throttle angle to reduce the speed error. The great torque demand is accompanied by large engine fuel consumption in Fig. 17(c). In Fig. 17(b), (c) and (e), compared with heuristic initialization, ECMS initialization shows better alignment with heuristic warm start final result, which is reflected by the cumulative fuel consumption in Fig. 16(d).

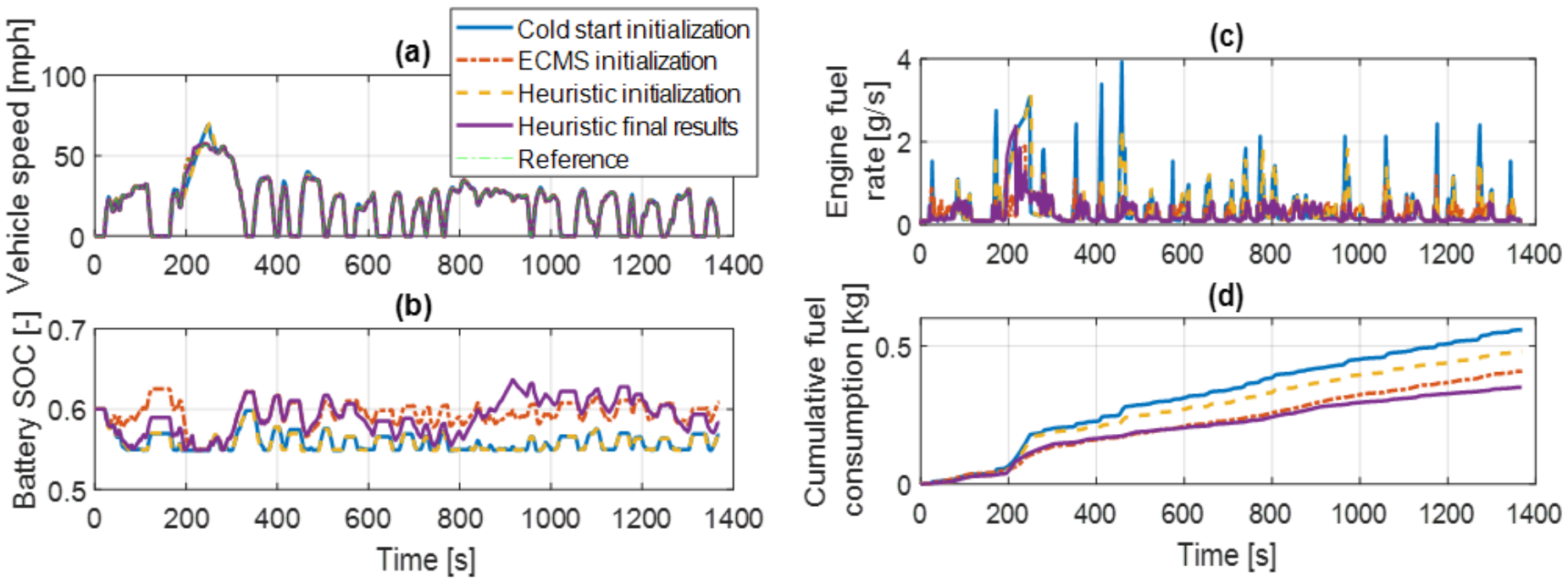

Fig. 16. Comparison of Q-learning simulation results among cold start initialization, ECMS warm start initialization, heuristic warm start initialization and heuristic warm start final the results after 5000 iterations: (a) vehicle speed (the reference is the driving cycle speed target), (b) battery SOC, (c) Engine fuel rate and (d) cumulative fuel consumption. (Initialization means the first iteration of the 5000 iterations with optimal policy and no random action, final results means the last iteration of the 5000 iterations with optimal policy and no random action.)

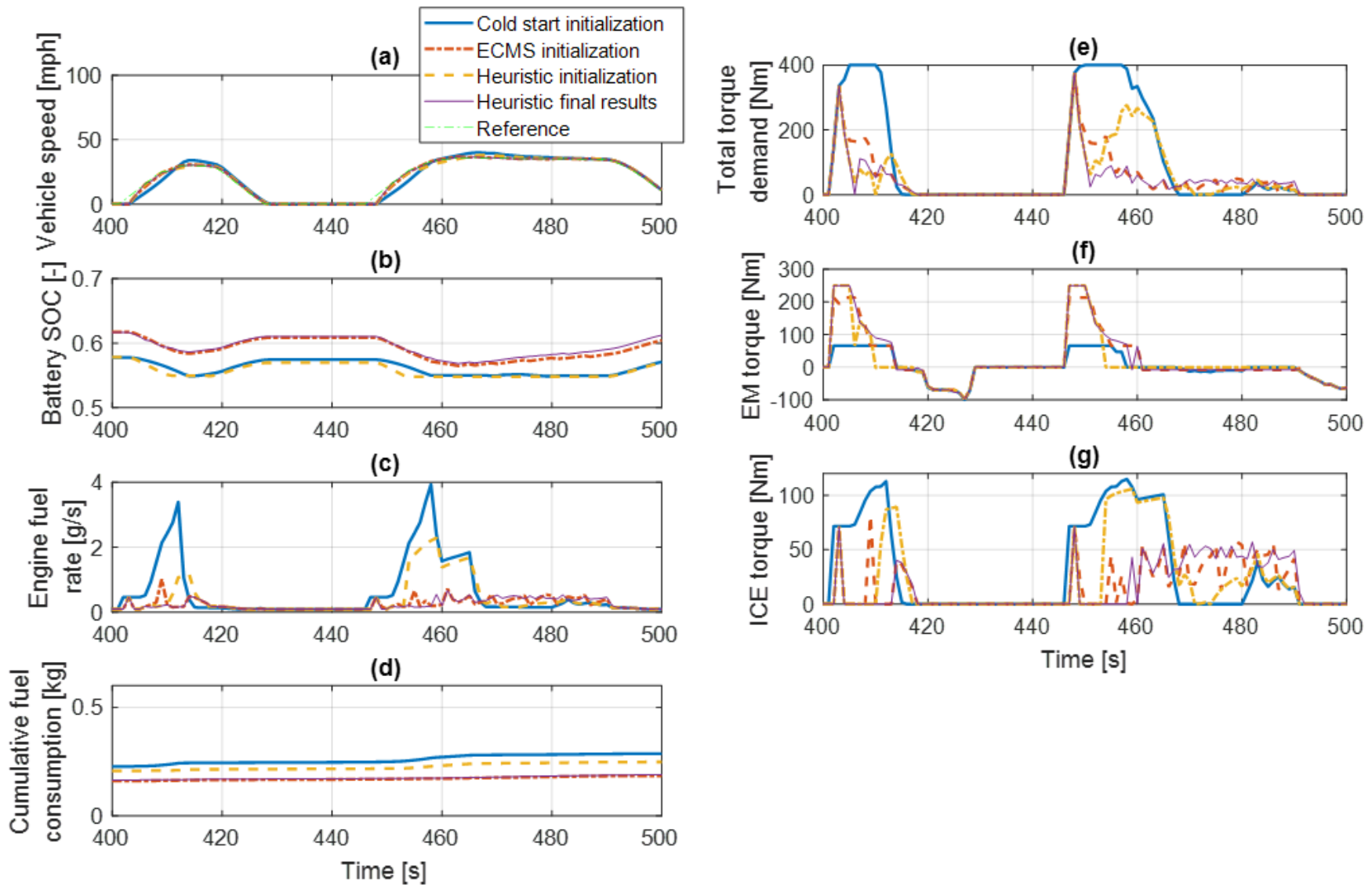

Fig. 17. Comparison of Q-learning simulation results among cold start initialization, ECMS warm start initialization, heuristic warm start initialization and heuristic warm start final the results after 5000 iterations: (a) vehicle speed (the reference is the driving cycle speed target), (b) battery SOC, (c) Engine fuel rate, (d) cumulative fuel consumption, (e) total torque demand, (f) EM torque and (g) engine torque. (Initialization means the first iteration of the 5000 iterations with optimal policy and no random action, final results means the last iteration of the 5000 iterations with optimal policy and no random action)

### 4.3 *Comparison with ECMS*

In this subsection, heuristic warm start Q-learning is first compared with ECMS in the UDDS driving cycle, which aims to show the proposed warm start control with respect to existing supervisory control. After the comparison of supervisory controls, the trained Q-learning is then directly implemented in two different driving cycles, which aims to validate the generality of the UDDS trained Q-learning. In other words, the Q value updates are only conducted in the UDDS driving cycle, and there is no Q value update in the two validation driving cycle simulation. The $\varepsilon$ is set as 0 in the $\varepsilon$-greedy action exploration method; thus the optimal policy is generated in the validation.

In the UDDS driving cycle simulation, the vehicle using ECMS supervisory control results in 54.6mpg. In comparison, the Q-learning with heuristic warm start results in 59mpg, which is around 10% fuel economy improvement. Simulation results comparison is shown in Fig. 18. The main difference between these two supervisory controls is the torque split in vehicle starting phases (i.e., 350s, 410s, and 450s in Fig. 18(a)). As shown in Fig. 18(f) and (g), ECMS uses more ICE than EM at vehicle starting phases, whereas Q-learning uses more EM than ICE. The ECMS supervisory control leads to a slightly greater vehicle speed tracking error, as shown in Fig. 18(a). The vehicle speed tracking error is then translated to great vehicle total torque demand as the virtual driver (i.e., PI controller) tries to reduce the speed gap. After three vehicle starting phases, the cumulative fuel consumption between ECMS and Q-learning reaches a gap. The difference in results reflects one intrinsic difference between ECMS and Q-learning, which is the model-based or model-free. In some existing publications [33, 34], backward-looking vehicle model is used. In the backward-looking vehicle model, the total vehicle torque demand or power demand is calculated using vehicle dynamics, assuming vehicle speed is perfectly matching driving cycle speed. In the simulations, the total vehicle torque demand is pre-known and fixed, which is not affected by different supervisory controls. However, in reality, the vehicle speed cannot perfectly match the reference speed as the vehicle operating conditions are impacted by the decision of supervisory controls. In this paper, a forward-looking vehicle model is used, which models the driver using a PI controller, and there is no perfect speed tracking assumption. This is why the total vehicle torque demand is different in Fig. 18(e). ECMS does not adapt to the different vehicle speed tracking performance, whereas Q-learning can adapt to this situation. More specifically, a slightly large vehicle tracking error is detrimental to fuel economy. Q-learning adapts its Q values to improve the fuel economy during the iterations. Though Q-learning is model-free, the vehicle tracking error results are reflected in the state and reward feedback from the vehicle (i.e., large speed error results in high total torque demand and large fuel consumption).

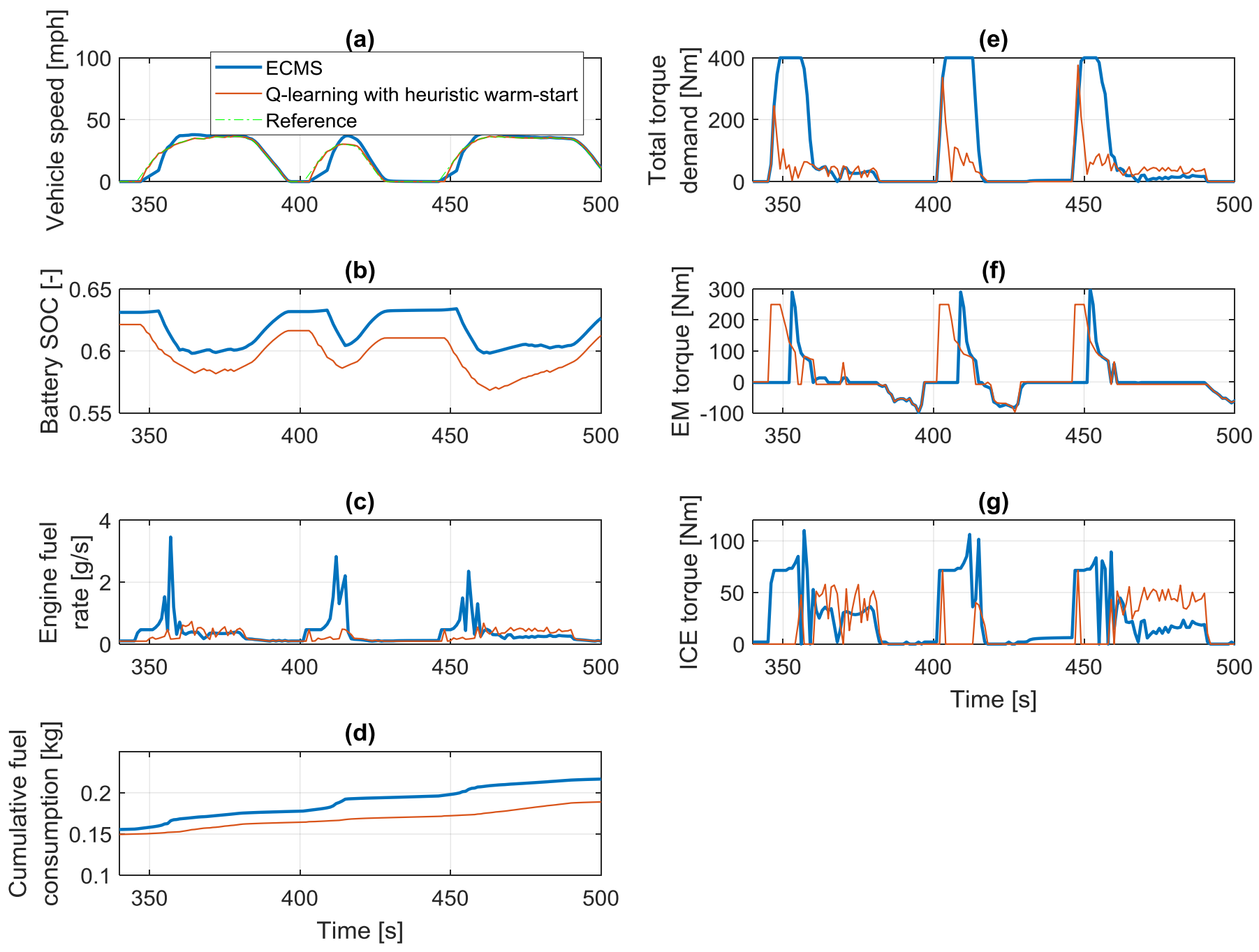

Fig. 18. Comparison of ECMS and Q-learning with heuristic warm start in UDDS driving cycle: (a) vehicle speed (the reference is the driving cycle speed target), (b) battery SOC, (c) Engine fuel rate, (d) cumulative fuel consumption, (e) total torque demand for vehicle, (f) EM torque and (g) engine torque.

To validate the learned Q-learning agent in different driving conditions, trained Q-learning from the UDDS driving cycle is then implemented in WLTP [35] and HWFET driving cycles [32]. ECMS is implemented as the comparison strategy, and the fuel economy results are shown in Table 4. In both driving cycles, the fuel economy from the Q-learning supervisory control shows higher values than that from the ECMS supervisory control. The fuel economy improvement is in the range of 10-16%, which is not far from the improvement (i.e., 10%) in the training driving cycle - UDDS. Thus, the training and validation results are consistent.

TABLE 4.

Fuel Economy of ECMS and Q-learning at different driving cycles.

(Note: Q-learning strategy is trained in UDDS driving cycle)

| Driving cycle | ECMS MPG | Q-learning MPG |
|---|---|---|
| WLTP | 36.6 | 42.5 |
| HWFET | 53.1 | 58.3 |

To implement the proposed warm start Q-learning supervisory control in an actual vehicle, the real-time computation feasibility is pivotal. In literature [14], a Q-learning control was implemented in an HEV. The experimental results showed that the Q-learning algorithm only took 0.447s CPU time to find the optimal action in the controller hardware at each time step. 0.5-1.0s time range is the controller sampling interval time in the actual vehicle controllers reported in [36, 37]. The 0.447s is shorter than controller sampling time, and thus the real-time execution is possible for the Q-learning supervisory control. Besides the real-time optimal action search in the Q values, the Q-learning needs to update the Q values based on the collected data during the daily vehicle operation. In actual vehicle deployment, the Q value update process is only executed when the vehicle is at rest, such as in the evening. The Q value update takes 1370 time steps in this study, which is corresponding to 612s (i.e., 10mins) if one time step takes 0.447s CPU time of the controller hardware. Based on the analysis of real-time execution and Q-learning update time, the proposed ensemble control has the potential in real-time implementation. The schematic of vehicle implementation is shown in Fig. 19. The Q values are first initialized in the controller software design. The data used in the initialization is generated either by actual vehicle operation over one driving cycle with ECMS control or by expert knowledge. As shown in Table 2, this process is described from line 4 to line 9. The software is then loaded into the controller hardware, which is followed by vehicle operation to collect experience for Q value update. When the vehicle is at rest, the controller hardware undergoes Q value update, which includes experience evaluation, Q value update and experience selection criteria update (see line 16-29 in Table 2). If the Q values converge (i.e., there is no significant change between the Q values prior to and after the update), the learning is finished. Otherwise, more experience will be collected during vehicle operation until Q values converge. If vehicle operation condition changes (e.g., vehicle owner moves from big city to remote urban area), the algorithm will detect the significant change of fuel economy. It then collects data and updates Q values until convergence.

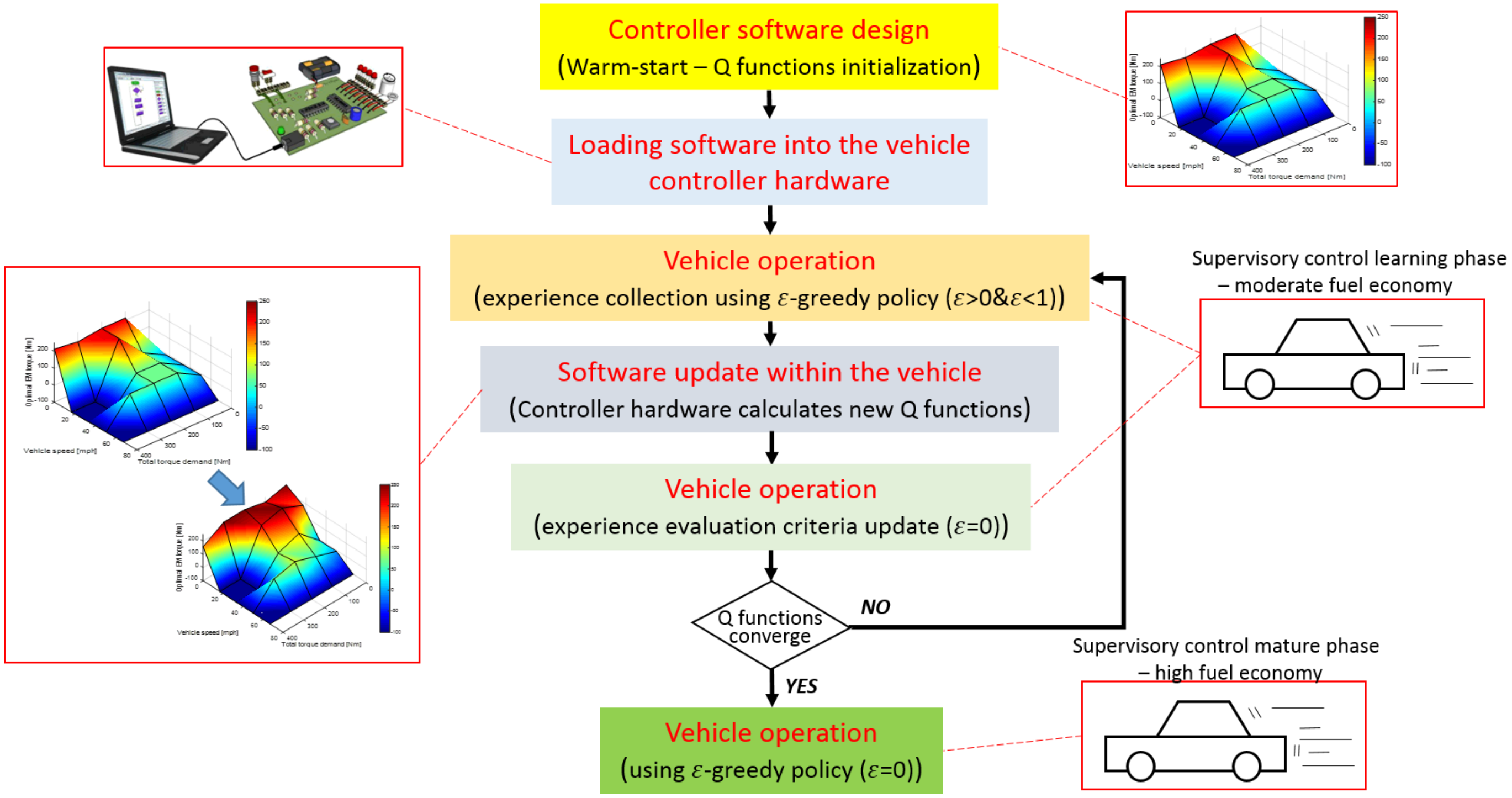

Fig. 19. Schematic of warm start Q-learning supervisory control vehicle implementation.

## V. Conclusion

This study proposed the warm start of RL-based supervisory control for a parallel HEV. ECMS and heuristic controls are used as warm start controls for the Q-learning supervisory control. Compared with cold start, both ECMS and heuristic warm start controls reduce the Q-learning iterations. The proposed ECMS warm start reduces the learning time as much as 68.8%. Right after the warm start Q value initialization, the ECMS and heuristic warm start Q-learning lead to 51mpg and 42mpg, compared with 36mpg from the cold start Q-learning. The proposed warm start methods significantly improve the overall fuel economy during the learning process. The warm start controls reduce the fuel consumption by reducing the engine torque saturation and allocating more torque to the EM. This torque distribution increases the operating efficiency of the engine. Compared with ECMS warm start, the heuristic warm start control has more flexibility in Q state-action value function initialization. Such a method appropriately transfers the learned knowledge to the Q-learning function. In the UDDS driving cycle that Q-learning iterations are conducted, the heuristic warm start Q-learning control shows 10% MPG higher than the ECMS control. In addition, the UDDS trained Q-learning control and ECMS control are compared in two different driving cycles for validation purposes. The results show 10-16% fuel saving by Q-learning compared with ECMS. The real-time implementation feasibility is discussed, and the vehicle implementation schematic is given as the guidance.

Even though substantial learning time reduction is achieved, there is room to improve. To reduce the RL warm start work, human behavior in the initial phase of learning can be studied. For example, rather than $\varepsilon$-greedy action exploration method, human action exploration is much more efficient and will be the focus of the next study. In addition, experimental work will help validate fuel economy and computation feasibility performance of the proposed supervisory control framework.